\documentclass[letterpaper, 10 pt, conference]{ieeeconf}  
\IEEEoverridecommandlockouts                   

\usepackage{graphics} 
\usepackage{epsfig}
\usepackage{mathptmx}
\usepackage{times} 
\usepackage{amsmath,amssymb,amsfonts}%
\usepackage{mathrsfs}%
\usepackage{outlines}
\usepackage{changes}
\usepackage{bm}
\usepackage{booktabs}
\usepackage{textcomp}
\usepackage{manyfoot}
\usepackage{multirow}
\usepackage{booktabs}
\usepackage{hyperref}
\hypersetup{
    colorlinks=true,
    linkcolor=blue,
    filecolor=magenta,      
    urlcolor=blue,
    pdftitle={Overleaf Example},
    pdfpagemode=FullScreen,
}
\title{\LARGE \bf
Bimanual In-hand Manipulation using Dual Limit Surfaces
}

\author{An Dang, James Lorenz, Xili Yi, Nima Fazeli% <-this % stops a space
\thanks{*This work was not supported by any organization}% <-this % stops a space
\thanks{University of Michigan, Ann Arbor, MI, USA
        {\tt\small <andang,jlorenz,yixili,nfz>@umich.edu}}%
}

\usepackage{todonotes}
\usepackage{xcolor}
\renewcommand{\vec}[1]{\bm{#1}}
\newcommand{\mat}[1]{\bold{#1}}

\begin{document}
\maketitle
\thispagestyle{empty}
\pagestyle{empty}

%%%%%%%%%%%%%%%%%%%%%%%%%%%%%%%%%%%%%%%%%%%%%%%%%%%%%%%%%%%%%%%%%%%%%%%%%%%%%%%%
\begin{abstract}
    In-hand object manipulation is an important capability for dexterous manipulation. In this paper, we introduce a modeling and planning framework for in-hand object reconfiguration, focusing on frictional patch contacts between the robot’s palms (or fingers) and the object. Our approach leverages two cooperative patch contacts on either side of the object to iteratively reposition it within the robot’s grasp by alternating between sliding and sticking motions. Unlike previous methods that rely on single-point contacts or restrictive assumptions on contact dynamics, our framework models the complex interaction of dual frictional patches, allowing for greater control over object motion. We develop a planning algorithm that computes feasible motions to reorient and re-grasp objects without causing unintended slippage. We demonstrate the effectiveness of our approach in simulation and real-world experiments, showing significant improvements in object stability and pose accuracy across various object geometries. For more information, visit \url{https://www.mmintlab.com/bimanual-dual-limit-surfaces/}.
\end{abstract}

%%%%%%%%%%%%%%%%%%%%%%%%%%%%%%%%%%%%%%%%%%%%%%%%%%%%%%%%%%%%%%%%%%%%%%%%%%%%%%%%
\section{INTRODUCTION}

In-hand manipulation refers to the adjustment of an object's pose relative to its end-effectors while it remains within the robot's grasp. Achieving this capability is crucial for dexterous robotic manipulation, where continuous control over the object is required to perform tasks such as regrasping, reorienting, or fine adjustment. Current robotic systems typically perform these tasks by either placing the object down and regrasping it or by using techniques such as finger gating. However, placing the object down requires precise knowledge of the environment and presents the risk of failing to regrasp the object in a manner suitable for subsequent tasks. Finger gating, meanwhile, often involves complex contact models or relies on unstable point contacts, limiting its practicality in real-world scenarios. These limitations highlight the need for more robust and flexible in-hand manipulation techniques that enable continuous control over the object without relinquishing the grasp.

Several approaches to in-hand manipulation have been explored in prior work, primarily focusing on model-based techniques that rely on environmental dexterous hands \cite{contact-rich-smooth, kurtz2023inverse, morgan2022complex, gao2021dexterous} or fixtures \cite{motion-cones,shi2017dynamic}. Environmental fixtures allow manipulation by using surrounding surfaces to guide the object's motion, but they require precise knowledge of the environment's frictional properties and often fail to fully exploit the dexterity of the robot's grippers. Anthropomorphic hands, on the other hand, typically rely on point contact models that require intricate reasoning about making and breaking contact during manipulation. This contact switching adds complexity to the planning process and can introduce instability, particularly in tasks that require continuous regrasping. These methods, while effective in certain scenarios, are limited by their dependence on either environmental constraints, restrictive contact models, or complex controls.

\begin{figure}[!t]\label{im:teaser}
\centering
\includegraphics[width=0.45\textwidth]{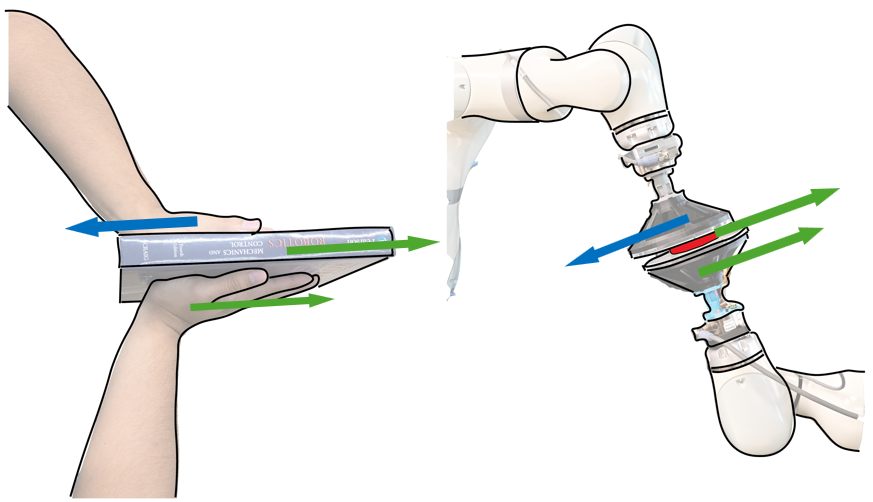}
\caption{In-hand manipulation of a grasped object using frictional patch contacts. The object maintains sticking contact with the lower palm, while maintaining sliding contact with the upper palm. As the lower palm moves, the object's position changes with respect to the upper palm, which enables object regrasping.}
\end{figure}

In this paper, we present a bimanual in-hand manipulation method that leverages dual frictional patch contacts made via palm end-effectors to regrasp objects without release or relying on external fixtures. The key feature of our method is the ability to control which surfaces are in sticking contact and which slide. This results in robust and precise object control, improving upon methods that rely solely on point contacts. Our approach builds upon the concept of dual limit surfaces \cite{kili}, which allow us to model the interaction between the object and the two patch contacts, reasoning about both sticking and slipping motions relative to each end-effector. To demonstrate the effectiveness of our approach, we use two 7-DoF Kuka IIWA Med arms, each equipped with custom palm end-effectors for patch contacts. We validate our method through both simulation and real-world experiments, showcasing its ability to regrasp objects with improved stability and accuracy across a variety of object geometries and manipulation tasks.

\section{Problem Statement}

% \begin{itemize}
%     \item Assumptions (not listed, can be organized in a sequence of sentences) -- object rigidity, Coulomb friction with patch contacts, and two robotic palms. 
%     \item Inputs to the problem -- object initial and final goals for example
%     \item Desired output -- end-effector motions that realize desired object motion.
% \end{itemize}

% Assumptions:
% \begin{itemize}
% \item Rigid object and end-effectors.
% \item Two robot palm end-effectors.
% \item Coulomb Friction. (2 patch contacts)
% \item Quasi-static motion.
% \end{itemize}

% Teaser figure
% put object on palm
% have reference frame
% one side opage + one side more transparent

% problem statement figure
% add in the labels
% include: patch contacts, q_o(t)
% important: highlight palm
% goal: show what it's supposed todo

%AN QUESTION: do we start off with this sentence?

We consider in-hand manipulation tasks where a bimanual robot must adjust the object pose relative to both of its palm end-effectors. We assume the robot has already grasped the rigid object and its grasp yields two patch contacts as seen in \ref{im:teaser}. We assume each patch contact obeys Coulomb friction law and any sliding motion is quasi-static (inertial forces are negligible). Let $\vec{q}_o^{\ell}(t), \vec{q}_o^r(t) \in \mathrm{SE}(2)$ denote the initial pose of the object w.r.t. the left and right palms respectively and $t \in [0, T]$ where $T$ is the total time. Given $\vec{q}_{des}^{\ell}, \vec{q}_{des}^{r} \in \mathrm{SE}(2)$, which are the desired relative object pose configurations, we wish to minimize the error between the desired object pose and the final relative poses of the object, expressed mathematically as:
% NOTE: still need to align s.t.
\begin{equation*}
\begin{aligned}
    \min_{\vec{q}_o^{\ell}, \vec{q}_o^r} \quad & \| \vec{q}_o^{\ell}(T) - \vec{q}_{des}^{\ell} \|_2^2 + \| \vec{q}_o^{r}(T) - \vec{q}_o^{r} \|_2^2
\end{aligned}
\end{equation*}
where $\vec{q}_o^{\ell *}(t)$ and $\vec{q}_o^{r *}(t)$ are the solutions we wish to find. The robot attempts to minimize this error by planning the motion of its palms in 3D space.

% This template provides authors with most of the formatting specifications needed for preparing electronic versions of their papers. All standard paper components have been specified for three reasons: (1) ease of use when formatting individual papers, (2) automatic compliance to electronic requirements that facilitate the concurrent or later production of electronic products, and (3) conformity of style throughout a conference proceedings. Margins, column widths, line spacing, and type styles are built-in; examples of the type styles are provided throughout this document and are identified in italic type, within parentheses, following the example. Some components, such as multi-leveled equations, graphics, and tables are not prescribed, although the various table text styles are provided. The formatter will need to create these components, incorporating the applicable criteria that follow.

\section{Related Works}

The problem of dexterous manipulation is difficult due to the high degrees-of-freedom and complex hand-to-environment contacts. With advances of machine learning, data-driven approaches are growing in popularity. AnyRotate \cite{yang2024anyrotate} uses reinforcement learning to rotate unseen objects, Morgan et al. \cite{morgan2022complex} uses sampling-based constrained multi-modal planning, and Hammoud et al. \cite{hammoud2024robotic} optimizes across a dictionary of learned manipulation primitives, to perform finger gaiting. Optimization methods \cite{sundaralingam2018geometric, khadivar2023adaptive} have demonstrated potential in finger gaiting for in-grasp manipulation, and classical control methods have been applied to the problem of in-hand object stabilization \cite{veiga2018hand, veiga2020grip}. These data-driven require longer computation times while our work has much shorter computation times due to its use of optimization. Additionally, these works rely on multiple fingers while our work focuses on only two for in-hand manipulation.
% \todo{AN->NIMA QUESTION: if we say it has shorter computation time, do we need a table for planning times?}

Modeling for contacts in robot manipulations are widely studied with long history. In \cite{howe1988sliding}, Howe et al. studied the contact between a robot finger and an object during sliding. To model the friction with patch contact, Goyal \cite{goyal-limit-surface} proposed limit surface, which is defined as the boundary of the set of all possible friction wrenches. Howe \cite{cutkosky-practical-limit-surfaces} proposed and compared different models for the limit surface, and showed a simplified ellipsoidal model can reduce computational complexity. In study \cite{xydas-soft-finger-limit-surface}, Xydas presented modeling and experimental results to evaluate the relationship between the normal force and the radius of contact for soft fingers. For \cite{motion-cones}, Nikhil developed a motion-cone planner to perform in-hand manipulation on a grasped object using environmental fixtures. Using friction cones and limit surfaces, they were able to formulate motion cones to drive the object to a given pose within grasp. For an planar object with both top and bottom contact, Yi et al. \cite{kili} derived a dual limit surfaces contact model, and planning algorithm for sliding on horizontal plane. In a subsequent work from Yi et al. \cite{kili_an}, they extended the contact model and sliding planning algorithm to tilted cases. In this study, we extend the dual limit surfaces model in tilted case into bimanual manipulation, and develop a planning algorithm based on this. 

% data driven approaches require lots of time to run
% dextrous manipulation methods mostly use point contact slack frictional contact patches, require more than two fingers, with higher degrees of freedom, and/or tactile feedback

% In this work, based on dual limit surface friction model, we use optimization-based planning algorithm to do bimanual manipulation to move an object in hand.

\section{Method}

\begin{figure*}[!t]\label{im:method}
\centering
\includegraphics[width=0.95\textwidth]{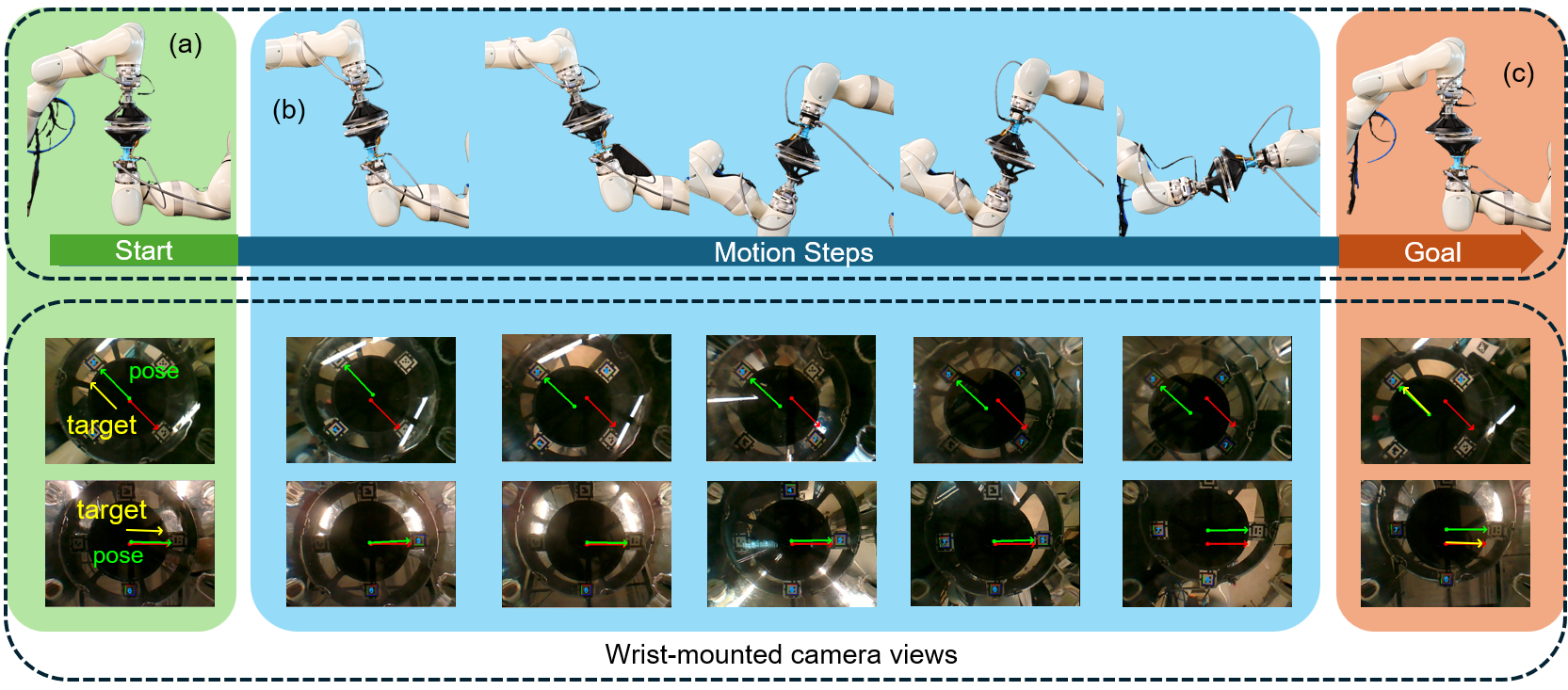}
\caption{Iterative repositioning of the grasped object by alternating sticking and sliding contact between the two robot palms.}
\end{figure*}

Our bimanual in-hand regrasp strategy is to drive the object to its target configuration by alternating each palm's movement while the opposing palm remains static. The core of our strategy is to plan for arm motions that result in the object maintaining sticking contact with the moving arm while sliding with respect to the static arm. We refer to this as slippage-free sliding. 
% This is a particularly effective strategy given that the object motion will be open-loop stable and the object will act as a rigid extension of the moving arm. 
Thus, we formulate our problem as:
\begin{equation*}
\begin{aligned}
    \min_{\vec{q}_o^{\ell}, \vec{q}_o^r} \quad & \| \vec{q}_o^{\ell}(T) - \vec{q}_{des}^{\ell} \|_2^2 + \| \vec{q}_o^{r}(T) - \vec{q}_o^{r} \|_2^2 \\
    \textrm{s.t.} \quad & g(\vec{q}_o^{\ell}, \vec{q}_o^{r}) < 0 
\end{aligned}
\end{equation*}
where we have introduced the slippage-free constraint. Our method is composed of a mechanics section and a planning section. The mechanics section focuses on deriving this constraint by assuming our two patch contacts can be modelled as dual limit surfaces \cite{kili}. The planning section provides an algorithm that solves the optimization program.

\subsection{Mechanics}

% while ensuring we follow motions that ensure no slippage $g(\vec{q}_o^{\ell}, \vec{q}_o^{r}) < 0$. We formulate our problem as: \textrm{s.t.} \quad & g(\vec{q}_o^{\ell}, \vec{q}_o^{r}) < 0
% In \todo{This sounds more like a method detail -- that we want to ensure slippage-free constraints in our formulation. There may be other approaches that don't make this assumption}order to successfully adjust our object pose for in-hand manipulation, we must ensure our slippage-free constraint $g(\vec{q}_o^{\ell}, \vec{q}_o^{r}) < 0$ is satisfied. We derive this constraint in a later section by assuming our two patch contacts can be modelled as dual limit surfaces \cite{kili}.

During our bimanual in-hand regrasp, we wish to characterize the resulting object motion due to alternating motions of the palms with respect to each other. Our goal is to derive a constraint for which the object maintains sticking contact with the moving palm while it slides with respect to the static palm. We refer to this as slippage-free sliding. In this section, we present the mechanics that describe this interaction (the dual limit surface model) and apply it to derive the slippage-free twist constraint for in-hand regrasping.

% While one of the palms is in motion, we wish to maintain sticking contact between the object and the moving palm. 
% The core of our bimanual in-hand regrasp strategy is to drive the object to its target configuration by alternating each palm's movement while the opposing palm remains static. While one of the palms is in motion, we wish to maintain sticking contact between the object and the moving palm. We refer to this as slippage-free sliding. In this section, we present the mechanics that describe this interaction (the dual limit surface model) and apply it to derive the slippage-free twist constraint for in-hand regrasping.

\noindent\subsubsection{\textbf{Limit Surface}}

The limit surface is a patch contact friction model which models the boundary of the set of all possible friction wrenches a patch contact can offer \cite{goyal-limit-surface}. While it is not tractable to calculate the closed-form equation representing the limit surface, we will use an ellipsoidal approximation, as presented by Howe and Cutkosky \cite{cutkosky-practical-limit-surfaces} and experimentally verified in \cite{xydas-soft-finger-limit-surface}. We denote $\vec{w} = \begin{bmatrix} f_x & f_y & m_z \end{bmatrix}^T$ as the friction wrench. Assuming isotropic friction, the ellipsoidal limit surface boundary $\vec{w}^T \mat{A} \vec{w} = 1$ can be described in terms of the friction coefficient $\mu$, normal force $N$, and the patch radius $r$ where $\mat{A}(\mu, N, r) = \text{Diag}\{(\mu N)^{-2}, (\mu N)^{-2}, (\mu c r N)^{-2}\}$. Assuming a uniform pressure distribution, we obtain $c = 0.6$ as shown in other works \cite{xydas-soft-finger-limit-surface, inhand-sliding}.

When an object is sliding on the patch contact, we can obtain the corresponding friction wrench, given an object twist expressed in the contact frame $\vec{v}_c = \begin{bmatrix} v_x, v_y, \omega_z \end{bmatrix}^T$. Using the result from \cite{inhand-sliding}, the maximum dissipation principle allows us to express the friction wrench $\vec{w}_c$ as:

\begin{equation}\label{eqn:twist2wrench-limit-surface}
    \vec{w}_c = -\frac{\mat{A}^{-1}\vec{v}_c}{\sqrt{\vec{v}_c^T \mat{A}^{-1} \vec{v}_c} }
\end{equation}

\noindent\subsubsection{\textbf{Dual Limit Surface for Object Tabletop Sliding}}
The dual limit surface model was originally developed for tabletop object sliding and considers two patch contacts that were modelled as limit surfaces. The ellipsoidal approximation allow us to analyze the contact mechanics and derive a slippage-free constraint. Let $\mat{A}(\mu_a, N_a, r_a)$ be the limit surface ellipsoid matrix for the end-effector/object contact while $\mat{B}(\mu_b, N_b, r_b)$ is the limit surface ellipsoid matrix for object/tabletop contact.

In order to ensure the object is sliding on the tabletop while the end-effector is sticking on the object, we impose constraints in the wrench space for the dual limit surfaces. We wish for the friction wrench $\vec{w}_a$ for the object/tabletop contact to stay on the limit surface boundary while the friction wrench for the end-effector/object $\vec{w}_b$ contact to stay within the limit surface boundary. These constraints are written as:
\begin{equation}\label{eqn:obj2table-slide}
    \vec{w}_a^T\mat{A}\vec{w}_a = 1
\end{equation}
\begin{equation}\label{eqn:obj2ee-stick}
    \vec{w}_b^T\mat{B}\vec{w}_b < 1
\end{equation}

% BIG NOTE: took from AR paper
When the tabletop is inclined, the gravity force $m\vec{g}$ decomposes into 2 components, where one acts as a net wrench offsetting the friction wrench from the limit surface while the other acts as an additive scalar to the normal force of the limit surface. We represent the gravity force in the contact frame as $\vec{g}_c = m\mat{J}_c\vec{g}$ where $\mat{J}_c$ is the contact jacobian. The vector $\vec{g}_f = \textit{Diag}\{1,1,0\}\vec{g}_c$
is the projection of the contact frame gravity into xy plane. Then $g_n = \begin{bmatrix} 0 & 0 & 1 \end{bmatrix} \vec{g}_c$ extracts the component of gravity vector that affects the normal force of the contact patch.

Assuming quasi-static motion, we can relate $\vec{w}_a$, $\vec{w}_b$, and $\vec{g}_f$ as follows:
\begin{equation}\label{eqn:quasi-static}
    \vec{w}_a + \vec{w}_b + \vec{g}_f = 0
\end{equation}
Additionally, we can relate the normal forces $N_a$ and $N_b$ with $N_a = N_b + g_n$.

Using Equation \ref{eqn:quasi-static}, we can solve for $\vec{w}_b$ and substitute it inside of Equation \ref{eqn:obj2ee-stick} to get.
\begin{equation}\label{eqn:og-tabletop-limit-surface-equation}
    \vec{w}_a \mat{B} \vec{w}_a + 2\vec{g}_f^T\mat{B}\vec{w}_a + \vec{g}_f^T \mat{B} \vec{g}_f < 1
\end{equation}
In order to integrate the equality constraint from Equation \ref{eqn:obj2table-slide}, we substitute the LHS of Equation \ref{eqn:obj2table-slide}, into the RHS of Equation \ref{eqn:og-tabletop-limit-surface-equation} and get the following:
\begin{equation}\label{eqn:wrench-constraint-tilted}
    \vec{w}_a^T(\mat{B} - \mat{A})\vec{w}_a + 2\vec{g}_f^T\mat{B}\vec{w}_a + \vec{g}_f^T\mat{B}\vec{g}_f < 0
\end{equation}
which gives us the slippage-free wrench constraint.

We convert this slippage-free wrench constraint into a twist constraint by using Equation \ref{eqn:twist2wrench-limit-surface} to substitute $\vec{w}_a$ with $\vec{v} \neq 0$ and get:
\begin{equation}\label{eqn:velocity-constraint-tilted}
\begin{aligned}
    \vec{v}^T (\mat{A}^{-1} \mat{B} \mat{A}^{-1} - \mat{A}^{-1}) \vec{v} - 2(\sqrt{ \vec{v}^T \mat{A}^{-1}\vec{v}}) \vec{v}^T \mat{A}^{-1} \mat{B} \vec{g}_f + \\ (\vec{g_f}^T\mat{B}\vec{g_f})\vec{v}^T\mat{A}^{-1} \vec{v} < 0
\end{aligned}
\end{equation}

In this tabletop case for \cite{kili}, it was found that Eq. \ref{eqn:velocity-constraint-tilted} was a 4\textsuperscript{th} order polynomial in terms of $N_b$. With a sufficiently large $N_b$, only enforcing the leading coefficient of the 4\textsuperscript{th} order polynomial to be negative was enough to ensure slippage-free sliding. This constraint was found to be:
\begin{equation}\label{eqn:leading-coefficient-4th-order}
    \vec{v}^T(\hat{\mat{A}}^{-1} \hat{\mat{B}}^{-1} \hat{\mat{A}}^{-1} - \hat{\mat{A}}^{-1}) \vec{v} < 0
\end{equation}
where $\hat{\mat{A}} = N_a^2\mat{A}$ and $\hat{\mat{B}} = N_b^2\mat{B}$.

% EXTRA CHUNK FOR THE TILTED TWIST STUFF
% For the tilted case in \cite{kili}, it was found that Eq. \ref{eqn:velocity-constraint-tilted} was a 4th-order polynomial in terms of $N_b$. If the leading coefficient of the 4\textsuperscript{th} order polynomial was negative, it was found that applying enough normal force $N_b$ would eventually lead to Equation \ref{eqn:velocity-constraint-tilted} being satisfied. This leading coefficient $\alpha$ was found to be:

% \begin{equation*}
%     \alpha = \vec{v}^T(\hat{\mat{A}}^{-1} \hat{\mat{B}}^{-1} \hat{\mat{A}}^{-1} - \hat{\mat{A}}^{-1}) \vec{v}
% \end{equation*}
% \begin{equation}\label{eqn:leading-coefficient-4th-order}
%     \alpha < 0
% \end{equation}
% This constraint was then integrated into the motion planning optimization problem.
%NOTE: talk about limitations
% Limitations I'm thinking of describing
% -> pushing too hard does not bode well when ur no longer on a tabletop (tabletop is actually another robot)
% -> pushing too hard saturates the Normal force stuff which is what we rely on in this paper

\noindent\subsubsection{\textbf{Dual Limit Surfaces for Bimanual Inhand Manipulation}}
For in-hand regrasping, when one palm is moving relative to a static palm, we wish to consider the moving palm to be the ``end-effector" while the static palm is the "table" in the tabletop example. This enables us to re-use the analysis derived in the tabletop case where $\mat{A}(\mu_a, N_a, r_a)$ is the limit surface ellipsoid matrix describing the contact between the static palm and the object while $\mat{B}(\mu_b, N_b, r_b)$ is the limit surface ellipsoid matrix describing the contact between the moving palm and the object. 

Additionally, we assume both palms have the same frictional properties with the object $\mu = \mu_a = \mu_b$, and the moving palm is always under the object $N_b = N_a + g_N$, rather than atop the object as in the tabletop case.

% For bimanual in-hand manipulation, we assume two robot manipulators with similar palm end-effectors. These palm end-effectors are grasping an object whose surface uniformly has the same frictional properties. For in-hand manipulation, we want the object to stick on one palm while sliding on the other palm which will be our "tabletop". We will alternate which palm becomes the tabletop in order perform regrasping for both end-effectors. This enables us to re-use the analysis derived in the tabletop case.

\begin{figure}[h]\label{im:nonconvex_constraint}
\centering
\includegraphics[width=0.35\textwidth]{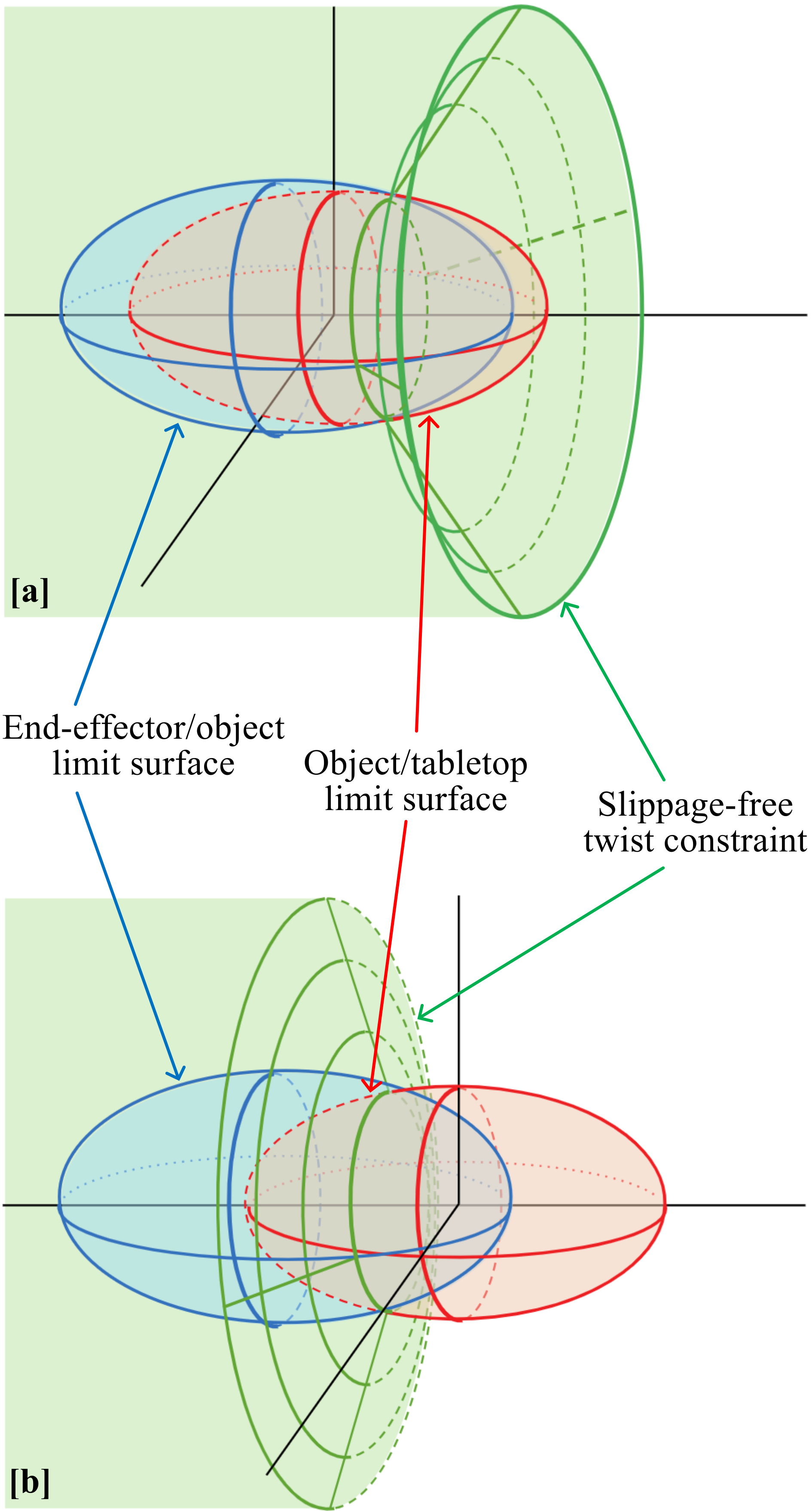}
\caption{Visualization of the wrench space for bimanual manipulation. The red ellipsoid represents $\vec{w}_a\mat{A}\vec{w}_a = 1$ while $\vec{w}_b\mat{B}\vec{w}_b = 1$ is the blue ellipsoid. The green region shows the slippage-free twist constraint. A non-convex slippage-free green region occurs when the blue ellipsoid covers over half of the red ellipsoid surface, shown in [a]. A convex slippage-free green region occurs when the blue ellipsoid covers less than half of the red ellipsoid surface, [b].}
\end{figure}

We first consider the case where the contact patch radius is the same $r = r_a = r_b$. This allows us to relate $\mat{A}$ to $\mat{B}$ through $\mat{B} = c\mat{A}$ where $c = \frac{N_a^2}{N_b^2}$, and simplifies Eq. \ref{eqn:velocity-constraint-tilted} into:
\begin{equation}\label{eqn:soc-simple}
    (c - 1 + c \vec{g}_f^T\mat{A}\vec{g}_f)\| \mat{A}^{-\frac{1}{2}} \vec{v} \|_2 < 2c\vec{g}_f^T \vec{v}
\end{equation}
where when $(c - 1 + c \vec{g}_f^T\mat{A}\vec{g}_f) > 0$ gives a second-order cone (SOC). This SOC can be best visualized by \ref{im:nonconvex_constraint}. However, when the constant isn't positive, this is no longer a second order cone which can be seen in \ref{im:nonconvex_constraint}. Additionally, this constraint has a square root term which do not play nice with optimization solvers. Instead, we utilize a different constraint. When $(c - 1 + c \vec{g}_f^T\mat{A}\vec{g}_f) < 0$, we can enforce $\vec{g}_f^T\vec{v} > 0$ which will also satisfy the same constraint.

% \begin{figure}[h]\label{im:convex_constraint}
% \centering
% \includegraphics[width=0.45\textwidth]{figs/convex_constraint.png}
% \caption{This figure visualizes the wrench space for bimanual manipulation. The red ellipsoid represents $\vec{w}_a\mat{A}\vec{w}_a = 1$ while $\vec{w}_b\mat{B}\vec{w}_b = 1$ is the blue ellipsoid. The green region shows the slippage-free twist constraint which is convex. }
% \end{figure}

We now consider the more general case where $r_a \neq r_b$. We cannot break down Eq. \ref{eqn:velocity-constraint-tilted}, but we can decompose the constraint into two smaller constraints:
\begin{equation}\label{eqn:soc-uneven-og-constraint}
    \vec{v}^T(\mat{A}^{-1}\mat{B}\mat{A}^{-1} - \mat{A}^{-1})\vec{v} < 0
\end{equation}
\begin{equation}\label{eqn:soc-uneven-radius-unsimplified}
    -2(\sqrt{\vec{v}^T\mat{A}^{-1}\vec{v}})\vec{v}^T\mat{A}^{-1}\mat{B}\mat{g}_f + (\vec{g}_f^T\mat{B}\vec{g}_f)\vec{v}^T\mat{A}^{-1}\vec{v} < 0
\end{equation}
Satisfying these constraints would also lead to satisfying Eq. $\ref{eqn:velocity-constraint-tilted}$. We can further simplify Eq. \ref{eqn:soc-uneven-radius-unsimplified} into 
\begin{equation}\label{eqn:soc-uneven-radius}
    \| \mat{A}^{-\frac{1}{2}} \vec{v} \|_2 < \frac{2}{(\vec{g}_f^T\mat{B}\vec{g}_f)}\vec{g}_f^T\mat{A}^{-1}\mat{B}\vec{v}
\end{equation}
which is a SOC. With the slippage-free twist constraints obtained, we can apply these as constraints for our constrained optimization algorithm in the next section.

\subsection{Planning}
We formulate our motion planning problem as a constrained optimization problem which alternates between which palm maintains sticking contact with the object. We formulate a quadratic cost between the object's relative pose to the left palm $\vec{x}^L$ and its goal pose $\vec{x}^L_G$ and a quadratic cost between the object's relative pose to the right palm $\vec{x}^R$ and its goal pose $\vec{x}^R_G$. For our palm setup, we introduce a 2-norm constraint $\|x\|_Q^2 = x^TQx$ where $Q = \textrm{Diag\{1,1,0\}}$ to keep the object between each palm radius $r$ which is a convex quadratic constraint. We denote $\vec{v}$ to be the object velocity in the contact frame of the patch contact between a chosen palm and the object. We formulate constraints that move the object poses based on our stick/slip contact assumptions and enforce our dual limit surface slippage-free twist constraints. 

We formulate the optimization problem as follows:
\begin{equation*}
\begin{aligned}
    \min_{\vec{x}^L, \vec{x}^R, \vec{x}^O, \vec{v}} \quad & \sum_{t=1}^N \| \vec{x}_t^R - \vec{x}^R_G \|_2^2 + \| \vec{x}_t^L - \vec{x}^L_G \|_2^2 \\
    \textrm{s.t.} 
    \quad & \vec{x}^L_{t+1} = \begin{cases}
        \vec{x}^L_t & \textrm{$t$ is even}\\
        \vec{x}^L_t + R_z(\theta^L_t)\vec{v}_t & \textrm{$t$ is odd} \\
    \end{cases} \\
    \quad & \theta^L_{t+1} = \begin{cases}
        \theta^L_t & \textrm{$t$ is even} \\
        \theta^L_t + \begin{bmatrix} 0 & 0 & 1 \end{bmatrix}\vec{v}_t & \textrm{$t$ is odd}
    \end{cases} \\
    \quad & \vec{x}^R_{t+1} = \begin{cases}
        \vec{x}^R_t & \textrm{$t$ is odd}\\
        \vec{x}^R_t + R_z(\theta^R_t)\vec{v}_t & \textrm{$t$ is even} \\
    \end{cases} \\
    \quad & \theta^R_{t+1} = \begin{cases}
        \theta^R_t & \textrm{$t$ is odd} \\
        \theta^R_t + \begin{bmatrix} 0 & 0 & 1 \end{bmatrix}\vec{v}_t & \textrm{$t$ is even}
    \end{cases} \\
    \quad & (c - 1 + c \vec{g}_f^T\mat{A}\vec{g}_f)\| \mat{A}^{-\frac{1}{2}} \vec{v}_t \|_2 < 2c\vec{g}_f^T \vec{v}_t\\
    \quad & \vec{x}^L_N = \vec{x}^L_G \quad \vec{x}^R_N = \vec{x}^R_G \quad
    \forall t \in \{0,...,N\}
\end{aligned}    
\end{equation*}

% \begin{equation*}
%     \begin{aligned}
%     \min_{\vec{x}^L, \vec{x}^R, \vec{x}^O, \vec{v}} \quad & \sum_{t=1}^N \| \vec{x}_t^R - \vec{x}^R_G \|_2^2 + \| \vec{x}_t^L - \vec{x}^L_G \|_2^2 \\
%     \textrm{s.t.} \quad & \vec{x}^O_{t+1} = \vec{x}^O_{t} + \vec{v}_t \\
%     \quad & \vec{x}^L_{t+1} = 
%     \begin{cases}
%         \vec{x}^L_{t} + \vec{v}_t & \textrm{$t$ is even} \\
%         R_z(\vec{v}_t)(\vec{x}^L_{t}) & \textrm{$t$ is odd}
%     \end{cases} \\
%     \quad & \vec{x}^R_{t+1} = 
%     \begin{cases}
%         \vec{x}^R_t + \vec{v}_t & \textrm{$t$ is odd} \\
%         R_z(\vec{v}_t)(\vec{x}^R_{t}) & \textrm{$t$ is even}\\
%     \end{cases} \\
%     \quad & \| \vec{x}^L_t \|_Q^2 \leq r^2 \\
%     \quad & \| \vec{x}^R_t \|_Q^2 \leq r^2 \\
%     \quad & Eqn. (\ref{eqn:soc-simple}) \quad \textrm{or} \quad Eqn. (\ref{eqn:soc-uneven-og-constraint}),(\ref{eqn:soc-uneven-radius})
%     \end{aligned}
% \end{equation*}
% where $\forall t \vec{x}^L_t, \vec{x}^R_t, \vec{x}^O_t, \vec{v}_t \in SE(2)$
% After solving the optimization, we use $\vec{x}^L,\vec{x}^R,\vec{x}^O$ to drive the bimanual robot arms to drive the object to the desired poses.

\section{Experiments}

In this section, we evaluate our method against a baseline over a distribution of relative object goal poses, grasp inclines, and two object geometries. We perform expeiments in both simulation and the real-world. % run two versions of our algorithms: open-loop in simulation, and closed-loop in real-world. The closed-loop algorithm measures the object pose relative to each hand to inform the planner. Our validation is performed in both simulation and the real-world for different object geometries. 
We show our planner's ability to maintain sticking contact and thus drive the object to its goal poses while the baseline planner fails to drive the object due to unintended object slip.

\subsection{Simulation Setup}
\begin{figure}[h]
    \vspace{5mm}
    \centering
    \includegraphics[width=0.35\textwidth]{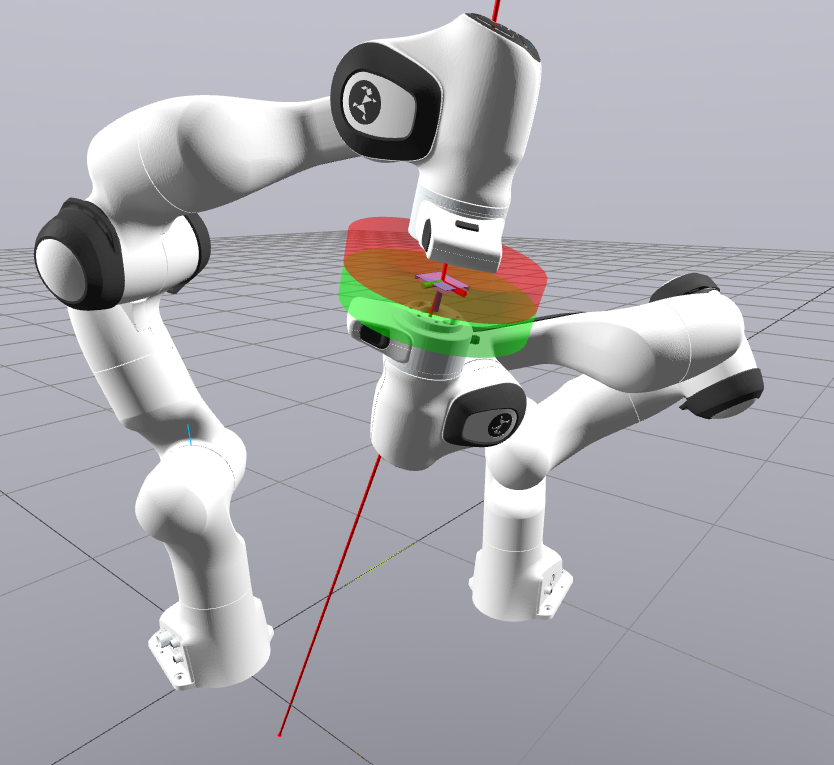}
    \caption{Simulated bimanual manipulation setup using Franka Emika Panda Robot Arms in Drake. We model patch contacts with the object using hydroelastic contacts, shown in red and green.}
    \label{im:sim-setup}
\end{figure}

We simulate our real-world setup using two Franka Emika Panda Robot Arms in Drake \cite{drake}, Fig.~\ref{im:sim-setup}. We use hydroelastic contacts to simulate the patch contacts, similar to \cite{kili}. We choose the same object geometries as the real-world setup, but additionally validate on objects with uneven contact surfaces (a smaller object surface on one side). 

\subsection{Real-World Experimental Setup} \label{sec:real-world-experimental-setup}
\begin{figure}[h]
    \centering
    \includegraphics[width=0.345\textwidth]{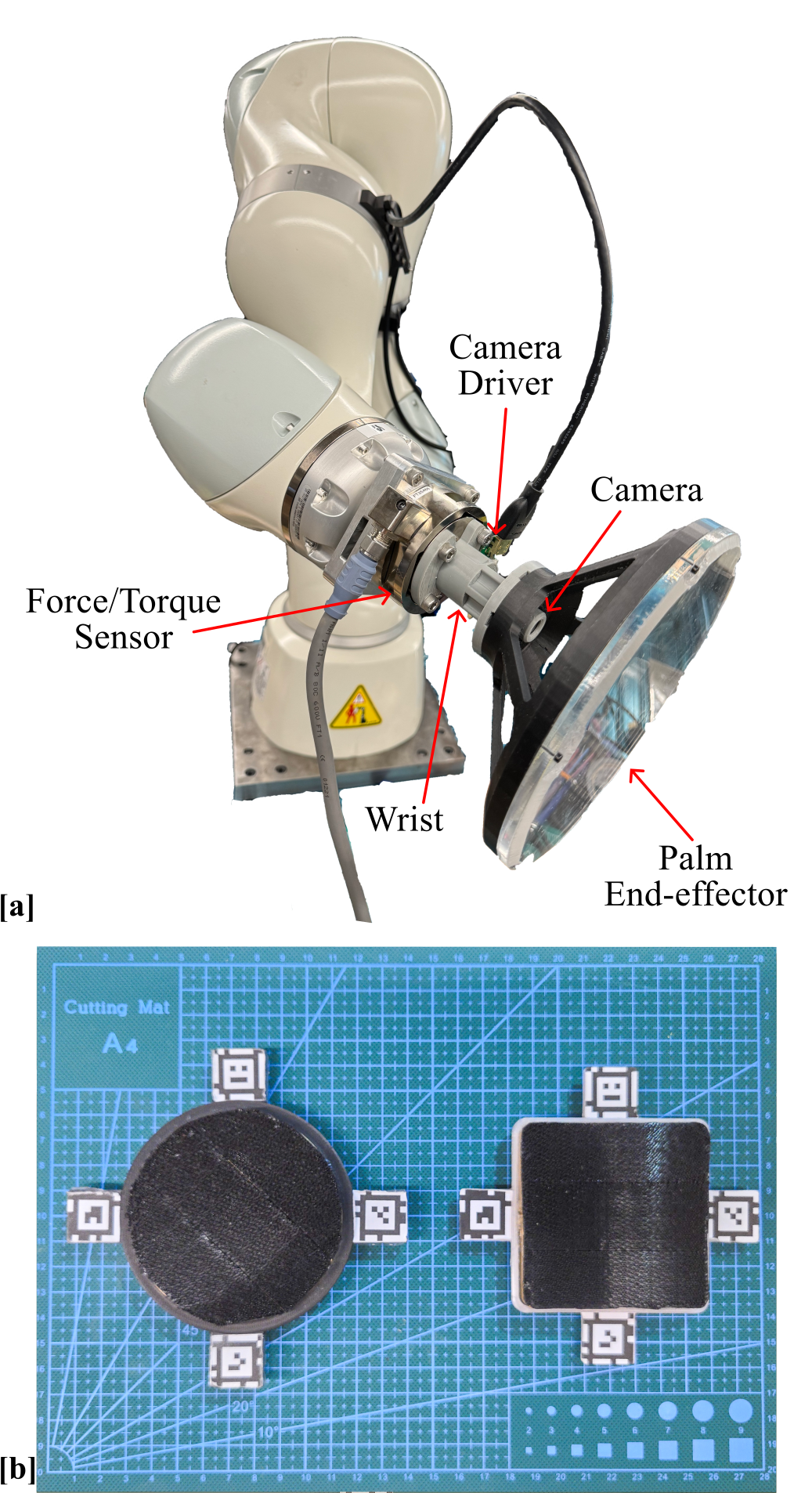}
    \caption{Real-world experimental setup: 7 DOF robot arm with a 6 DOF force/torque sensor, a transparent palm end-effector, and a camera in the palm to detect object pose, shown in [a]. Two object geometries -- square, circle -- each made of brass with self-engaging tape on the surface, [b].}
    \label{im:real-world setup}
\end{figure}

For the real-world setup, we use two Kuka MED LBR 14 R820 7 DOF robot arms each equipped with an ATI Gamma 6 DOF Force/Torque sensor to measure any external wrenches applied to each robot end-effectors. Each robot arm is equipped with a palm end-effector. These palms have a circular clear acrylic sheet with a clear polyurethane cover mounted on top. Inside each palm, there is a camera that sees through the acrylic and polyurethane to detect the object pose. Our robot setup can be seen in Fig.~\ref{im:real-world setup}.

For our object setup, we use brass plates of various geometries with self-engaging tape adhered to both sides. We chose our objects to be square and round discs. Each of these objects have the same thickness and mass. We mount 4 AprilTags \cite{apriltags} on each object for object pose detection. Our real-world objects are shown in Fig.~\ref{im:real-world setup}[b].

% \begin{figure}[h]
%     \centering
%     \includegraphics[width=0.45\textwidth]{figs/an-realworld.png}
%     \caption{Real-world experimental object setup}
%     \label{im:real-world-obj-setup}
% \end{figure}

\section{Results}
In this section, we show our results for planning in simulation and the real-world. 

% \begin{table}[]
% \centering
% \begin{tabular}{ccccc}
%  & \multicolumn{2}{c}{\textbf{obj2left}} & \multicolumn{2}{c}{\textbf{obj2right}} \\ \cline{2-5}
% \multicolumn{1}{l}{} & \multicolumn{1}{l}{\textit{\textbf{MSE}}} & \multicolumn{1}{l}{\textit{\textbf{STDEV}}} & \multicolumn{1}{l}{\textit{\textbf{MSE}}} & \multicolumn{1}{l}{\textit{\textbf{STDEV}}} \\ \hline
% \textbf{naive closed-loop} & 0.0 & 0.0 & 0.0 & 0.0 \\
% \textbf{naive open-loop} & 0.0 & 0.0 & 0.0 & 0.0 \\
% \textbf{ours closed-loop} & 0.0 & 0.0 & 0.0 & 0.0 \\
% \textbf{ours open-loop} & 0.0 & 0.0 & 0.0 & 0.0
% \end{tabular}
% \end{table}

%NOTE: make 2-col wide on page. Maybe try radius of gyration. Calc for each object.
\begin{table*}[] 
\vspace{10mm}
\centering
\caption{Real-world and simulation results.}
\begin{tabular}{l|cccc|cccc}
\centering
 & \multicolumn{4}{c|}{\textbf{Object Top}} & \multicolumn{4}{c}{\textbf{Object Bottom}} \\ \hline
\textbf{} & \textbf{RMSE} & \textbf{STDEV} & \textbf{RMSE} & \textbf{STDEV} &
\textbf{RMSE} & \textbf{STDEV} & \textbf{RMSE} & \textbf{STDEV} \\ 
\textbf{} & \textbf{(mm)} & \textbf{(mm)} & \textbf{(deg)} & \textbf{(deg)} &
\textbf{(mm)} & \textbf{(mm)} & \textbf{(deg)} & \textbf{(deg)} \\ 
\hline
\textbf{Circle, Simulation (Baseline)} & 24.98 & 18.30 & 54.02 & 36.67 & 27.12 & 19.96 & 54.30 & 36.82\\
\textbf{Circle, Simulation (Ours)} & 8.96 & 4.71 & 3.78 & 2.30 & 14.14 & 10.13 & 3.80 & 2.08\\
\textbf{Square, Simulation (Baseline)} & 24.96 & 18.25 & 54.22 & 36.85 & 27.13 & 19.95 & 55.05 & 37.47\\
\textbf{Square, Simulation (Ours)} & 9.07 & 4.81 & 3.30 & 2.01 & 14.34 & 10.32 & 3.33 & 1.8\\
\textbf{Circle, Real-World (Baseline)} & 27.33 & 16.94 & 53.64 & 34.27 & 23.11 & 13.92 & 75.80 & 50.07\\
\textbf{Circle, Real-World (Ours)} & 4.80 & 2.78 & 4.59 & 3.32 & 6.69 & 2.49 & 4.37 & 3.62\\
\textbf{Square, Real-World (Baseline)} & 22.06 & 10.36 & 58.29 & 41.69 & 18.18 & 7.74 & 52.99 & 38.64\\
\textbf{Square, Real-World (Ours)} & 5.61 & 3.13 & 2.01 & 1.26 & 6.35 & 2.82 & 1.42 & 0.72\\
\end{tabular}
\label{tab:real-world-results}
\end{table*}

% \begin{table}[]
% \centering
% \begin{tabular}{l|cc|cc}
% \toprule
%  & \multicolumn{2}{c|}{\textbf{obj2left}} & \multicolumn{2}{c}{\textbf{obj2right}} \\ 
% \cmidrule(r){2-3} \cmidrule(l){4-5}
% \textbf{} & \textbf{MSE} & \textbf{STDEV} & \textbf{MSE} & \textbf{STDEV} \\ 
% \midrule
% \textbf{naive closed-loop} & 0.0 & 0.0 & 0.0 & 0.0 \\
% \textbf{naive open-loop} & 0.0 & 0.0 & 0.0 & 0.0 \\
% \textbf{ours closed-loop} & 0.0 & 0.0 & 0.0 & 0.0 \\
% \textbf{ours open-loop} & 0.0 & 0.0 & 0.0 & 0.0 \\
% \bottomrule
% \end{tabular}
% \end{table}

\subsection{Simulation Results}\label{simulation-results}

\begin{figure}\label{fig:exemplar_result}
    \centering
    \includegraphics[width=0.6\linewidth]{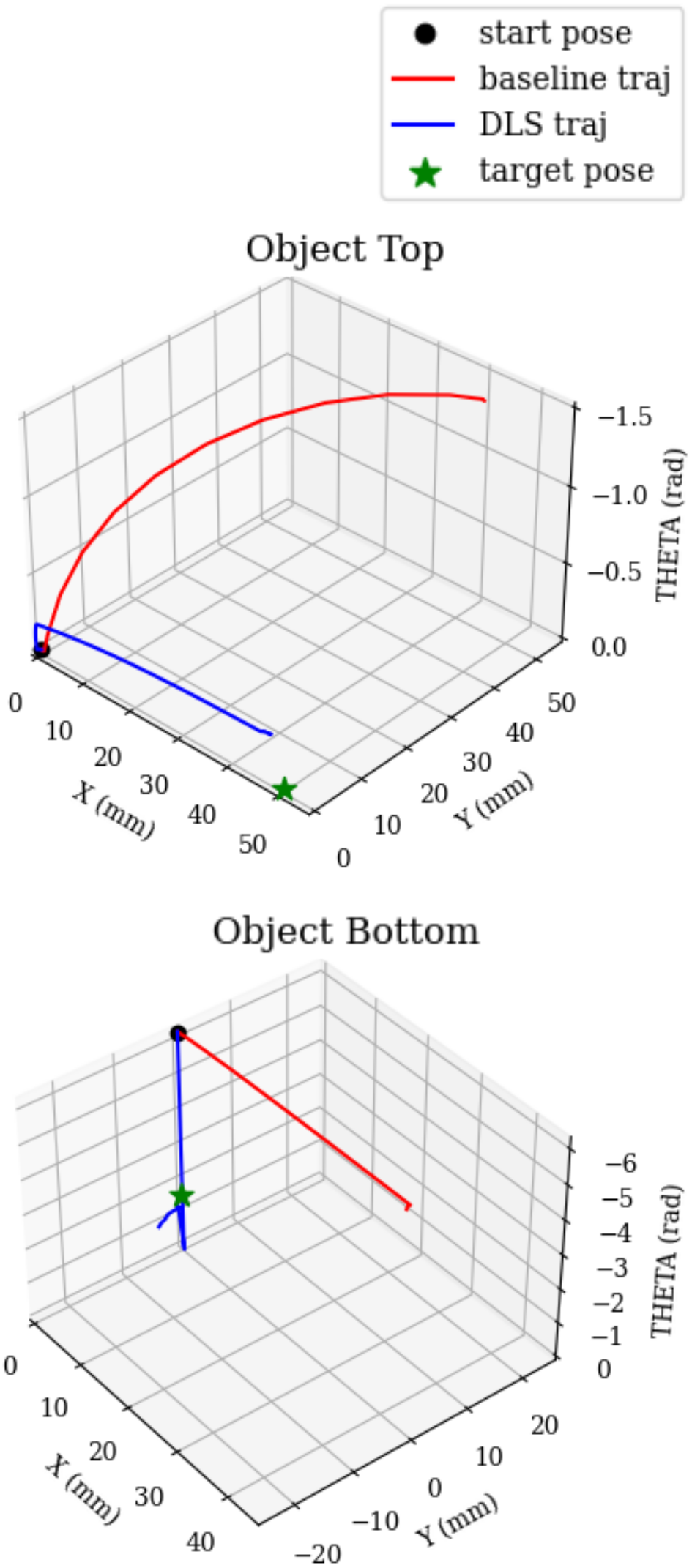}
    \caption{This is an exemplar plot of the trajectories collected from running our algorithm vs. the baseline in simulation. The baseline trajectory is marked in red while our algorithm's trajectory is marked in blue. The goal pose is denoted as a green star. }
    \label{fig:enter-label}
\end{figure}

Using the experimental setup described in Section \ref{sec:real-world-experimental-setup}, we manipulate the object to a series of goal poses within the grasp, creating a path along the palm. We tested on 3 separate paths, each with a unique set of goal poses. Each path was tested at 3 different angles of grasp incline: 20, 30, and 45 degrees from horizontal. The baseline attempts to move the object to each goal pose in a straight-line trajectory. In contrast, our algorithm plans a trajectory to each goal pose that maintains sticking contact between the grasped object and the moving palm. We tested across these conditions for each of the 3 grasped objects. To quantify the success of each experiment, we compare the error between the object's goal pose and measured pose. We averaged the results across all paths and angles, and report the results in Table \ref{tab:real-world-results} below. An exemplar path comparing our baseline trajectory and algorithm's trajectory is shown in \ref{fig:exemplar_result}.

Our algorithm results strongly outperformed the baseline; however, our algorithm did accumulate errors in both translation and rotation, likely due to the open-loop framework. When the object is no longer at the center of the palm, changing the in-hand pose requires rotating the palm about the object. As a result, even a small devation from the object's nominal position, the rotation of the palm about the object could cause large errors. This explains why we observe over 5 mm of translation error in our results. In the open-loop framework, accumulated small errors early in the path can amplify positional and translational errors later in the path. This observation motivates closing the loop for our real-world results.

\subsection{Real-World Results} \label{real-world-results}

To evaluate our planner against the baseline in simulation, we used the same set of paths and grasp angles as described in Section \ref{simulation-results}. Again, we measure success as the error between the object's goal posed and measured pose, and average our results across all paths and angles, shown in Table \ref{tab:real-world-results}. Unlike the simulation experiments, the real-world experiments were performed in closed-loop using object position for both our method and the baseline, to prevent accumulation of small errors in robot and measurement precision.

For both the square and circle objects, our algorithm outperformed the baseline in both translational and rotational error. While the robot executed the baseline, we observed that the object would lose sticking contact with the moving palm and slide in the direction of gravity, causing high translational and rotational error, particularly at steeper angles. Our algorithm reduced the frequency and magnitude of slipping between the moving palm and grasped object. We also observed that our algorithm produced a smaller standard deviation of error across trials, indicating an improvement in repeatability over the baselines. 

% \todo[]{@An pls improve this paragraph}However, we still observed minor translational and rotational error with our algorithm, likely due to compliance in our end effectors that were not accounted for in our algorithm. We also noted measurement error due to small imprecision in the placement of the AprilTags on the object. These errors can be addressed with stiffer end effectors, but are satisfactory for this work. 

% \begin{table}
% \centering
% \begin{tabular}{ p{1.5cm} p{1.5cm} p{1.5cm} p{1.5cm} }
% \hline
% \multicolumn{4}{c}{\textbf{Angle RMSE (rad)}} \\ \hline
% \multirow{2}{*}{\textbf{$r_e$ (m)}} & \multicolumn{3}{c}{\textbf{$\mu_p$}} \\ \cline{2-4} 
%  & 0.125 & 0.15 & 0.175\\ \hline
% 0.015 & 0.152 & 0.076 & 0.045 \\
% 0.025 & 0.132 & 0.063 & 0.037 \\
% 0.035 & 0.116 & 0.055 & 0.031 \\ \hline
% \end{tabular}
% \caption{Incomplete table.}
% \label{tb:heatmap_zeroshot}
% \end{table}

\section{Discussion and Limitations}

In this paper, we demonstrate an approach for in-hand object manipulation using frictional patch contacts to produce alternating sticking and sliding motion. The high translational and angular error in the baseline demonstrates how integrating slippage-free constraints in planning can enable the bimanual robots to drive the object to its in-hand goal poses. However, there are a few limitations to this work. Due to the fact that we control the motion of the palms in end-effector space, our implementation of bimanual in-hand manipulation is limited to local planning, which presents a challenge for our trajectory optimization. One approach to mitigate this limitation is to implement an analytic solution such as \cite{cohn2024planning} to our inverse kinematics to perform global planning. Moreover, objects that are tall, narrow, or lack flat surfaces may be susceptible to tipping while moving in-hand, which could result in slippage due to uneven pressure distribution across the contact patch, or even loss of the patch contact. While the object geometries we selected for this paper are flat, wide, and relatively thin, and thus satisfy the assumptions of the dual limit surface model, we may need a framework to apply dual limit surfaces with less stable object geometries to prevent unexpected slippage. 

% \james{ limitations of the ik -- we're only playing with end-effector space, but we can easily integrate the https://ieeexplore.ieee.org/abstract/document/10610675 to improve the ik and do global planning instaed of local planning; dual limit surfaces assume that you're using flat, relatively thin objects that won't "torque out" and cause us to lose the contact patch; slippage can occur due to uneven pressure distribution along the contact patch}

One motivation for applying our dual limit surface method to less stable object geometries is to enable regrasping of a wider variety of object shapes and sizes. This can be achieved with improved estimation of the shape and pressure distribution across the palm/object contact patch. Future work will explore tactile perception of the contact patch so that we can regulate or reason about the object's pressure distribution, and improve upon the results demonstrated in this paper. Finally, much like the majority of other in-hand object manipulation approaches, we have assumed object rigidity. Future work on expanding the notion of dual limit surfaces to deformable objects can significantly expand the set of objects we can manipulate.

% \james{using larger objects, figure out how to apply dual limit surface method with less stable geometries of objects (tall, not flat), if we have a bubble/gelslim type sensing of the pressure distribution along the contact patch so that we can manage or reason about the pressure distribution}

% We denote the first limit surface between the left palm and object $\mat{A}=\text{Diag}\{(\mu_\ell N_\ell)^{-2}, (\mu_\ell N_\ell)^{-2}, (\mu_\ell c r_\ell N_\ell)^{-2}\}$ and the second limit surface between the right palm and object $\mat{B}=\text{Diag}\{(\mu_r N_r)^{-2}, (\mu_r N_r)^{-2}, (\mu_r c r_r N_r)^{-2}\}$. We respectively define $\mu$, $c$, and $N$ to be the friction coefficient, limit surface constant (as found in \cite{kili}), and normal force of the patch contact.
\bibliographystyle{ieeetr}
\bibliography{IEEEfull}

\end{document}